\def\year{2022}\relax
\documentclass[letterpaper]{article} % DO NOT CHANGE THIS
\usepackage{aaai22}  % DO NOT CHANGE THIS
\usepackage{times}  % DO NOT CHANGE THIS
\usepackage{helvet}  % DO NOT CHANGE THIS
\usepackage{courier}  % DO NOT CHANGE THIS
\usepackage[hyphens]{url}  % DO NOT CHANGE THIS
\usepackage{graphicx} % DO NOT CHANGE THIS
\usepackage{natbib}  % DO NOT CHANGE THIS AND DO NOT ADD ANY OPTIONS TO IT
\usepackage{caption} % DO NOT CHANGE THIS AND DO NOT ADD ANY OPTIONS TO IT
\DeclareCaptionStyle{ruled}{labelfont=normalfont,labelsep=colon,strut=off} % DO NOT CHANGE THIS
\usepackage{algorithm}
\usepackage{algorithmic}
\usepackage[utf8]{inputenc} % allow utf-8 input
\usepackage[T1]{fontenc}    % use 8-bit T1 fonts
\usepackage{hyperref}       % hyperlinks
\usepackage{url}            % simple URL typesetting
\usepackage{booktabs}       % professional-quality tables
\usepackage{amsfonts}       % blackboard math symbols
\usepackage{nicefrac}       % compact symbols for 1/2, etc.
\usepackage{microtype}      % microtypography
\usepackage{xcolor}         % colors
\usepackage{graphicx}
\usepackage{amsmath}
\usepackage{tabularx}
\usepackage{stfloats}
\usepackage{url}
\usepackage{times}
\usepackage{epsfig}
\usepackage{graphicx}
\usepackage{amsmath}
\usepackage{bm}
\usepackage{amssymb}
\usepackage{color}
\usepackage{mathtools,xparse}
\usepackage{enumitem}
\usepackage{multirow}
\usepackage{rotating}
\usepackage{url}            % simple URL typesetting
\usepackage{booktabs}       % professional-quality tables
\usepackage{amsfonts}       % blackboard math symbols
\usepackage{nicefrac}       % compact symbols for 1/2, etc.
\usepackage{microtype}      % microtypography
\usepackage{tabularx}
\usepackage{arydshln}
\usepackage{siunitx}

\usepackage{natbib}
\newcommand{\ttg}[1]{text#1$\to$#1graph}
\newcommand*\raiseup[1]{%
	\begingroup
	\setbox0\hbox{\scriptsize\strut #1}%
	\leavevmode
	\raise\dimexpr \ht\strutbox - \ht0\box0
	\endgroup
}
\newcommand{\rttg}{R\raiseup{\ttg{}}}

\usepackage{newfloat}
\usepackage{listings}
\floatstyle{ruled}
\newfloat{listing}{tb}{lst}{}
\floatname{listing}{Listing}
\usepackage{graphicx}
\usepackage{amsmath}
\usepackage{tabularx}
\title{Improving Scene Graph Classification by Exploiting Knowledge from Texts}
\author {
    {Sahand Sharifzadeh,\textsuperscript{\rm 1}\equalcontrib}
    {Sina Moayed Baharlou,\textsuperscript{\rm 1}\equalcontrib\thanks{S. M. Baharlou contributed to this project while he was a visiting
researcher at the Ludwig Maximilian University of Munich.}}
    Martin Schmitt,\textsuperscript{\rm 2}\\
    Hinrich Schütze,\textsuperscript{\rm 2}
    Volker Tresp \textsuperscript{\rm 1,3}
}
\affiliations {
    \textsuperscript{\rm 1} Department of Informatics, LMU Munich, Germany\\
    \textsuperscript{\rm 2} Center for Information and Language Processing (CIS), LMU Munich, Germany\\
    \textsuperscript{\rm 3} Siemens AG, Munich, Germany\\
    sahand.sharifzadeh@gmail.com, sina.baharlou@gmail.com
}
\begin{document}

\maketitle

\begin{abstract}
Training scene graph classification models requires a large amount of annotated image data. Meanwhile, scene graphs represent relational knowledge that can be modeled with symbolic data from texts or knowledge graphs. While image annotation demands extensive labor, collecting textual descriptions of natural scenes requires less effort. In this work, we investigate whether textual scene descriptions can substitute for annotated image data. To this end, we employ a scene graph classification framework that is trained not only from annotated images but also from symbolic data. In our architecture, the symbolic entities are first mapped to their correspondent image-grounded representations and then fed into the relational reasoning pipeline.
Even though a structured form of knowledge, such as the form in knowledge graphs, is not always available, we can generate it from unstructured texts using a transformer-based language model. We show that by fine-tuning the classification pipeline with the extracted knowledge from texts, we can achieve {\raise.12ex\hbox{$\scriptstyle\sim$}}8x more accurate results in scene graph classification, {\raise.12ex\hbox{$\scriptstyle\sim$}}3x in object classification, and {\raise.12ex\hbox{$\scriptstyle\sim$}}1.5x in predicate classification, compared to the supervised baselines with only 1\% of the annotated images.
\end{abstract}

%Introduction
\section{Introduction}
\label{introduction}
Relational reasoning is one of the essential components of intelligence; humans explore their environment by grasping the entire context of a scene rather than studying each item in isolation from the others. Furthermore, we expand our understanding of the world by educating ourselves about novel facts through reading or listening. For example, we might have never seen a ``cow wearing a dress'' but might have read about Hindu traditions of decorating cows. While we already have a robust visual system that can extract basic visual features such as edges and curves from a scene, the description of a ``cow wearing a dress'' refines our visual understanding of relations on an object level and enables us to recognize a dressed cow when seeing it. 

Relational reasoning is gaining growing popularity in the Computer Vision community and especially in the form of scene graph (SG) classification. The goal of SG classification is to classify objects and their relations in an image. One of the challenges in SG classification is collecting annotated image data. Most approaches in this domain rely on thousands of manually labeled and curated images. In this paper, we investigate whether the SG classification models can be fine-tuned from textual scene descriptions (similar to the ``dressed cow'' example above).

\begin{figure*}
	\begin{center}
		\includegraphics[width=.85\textwidth]{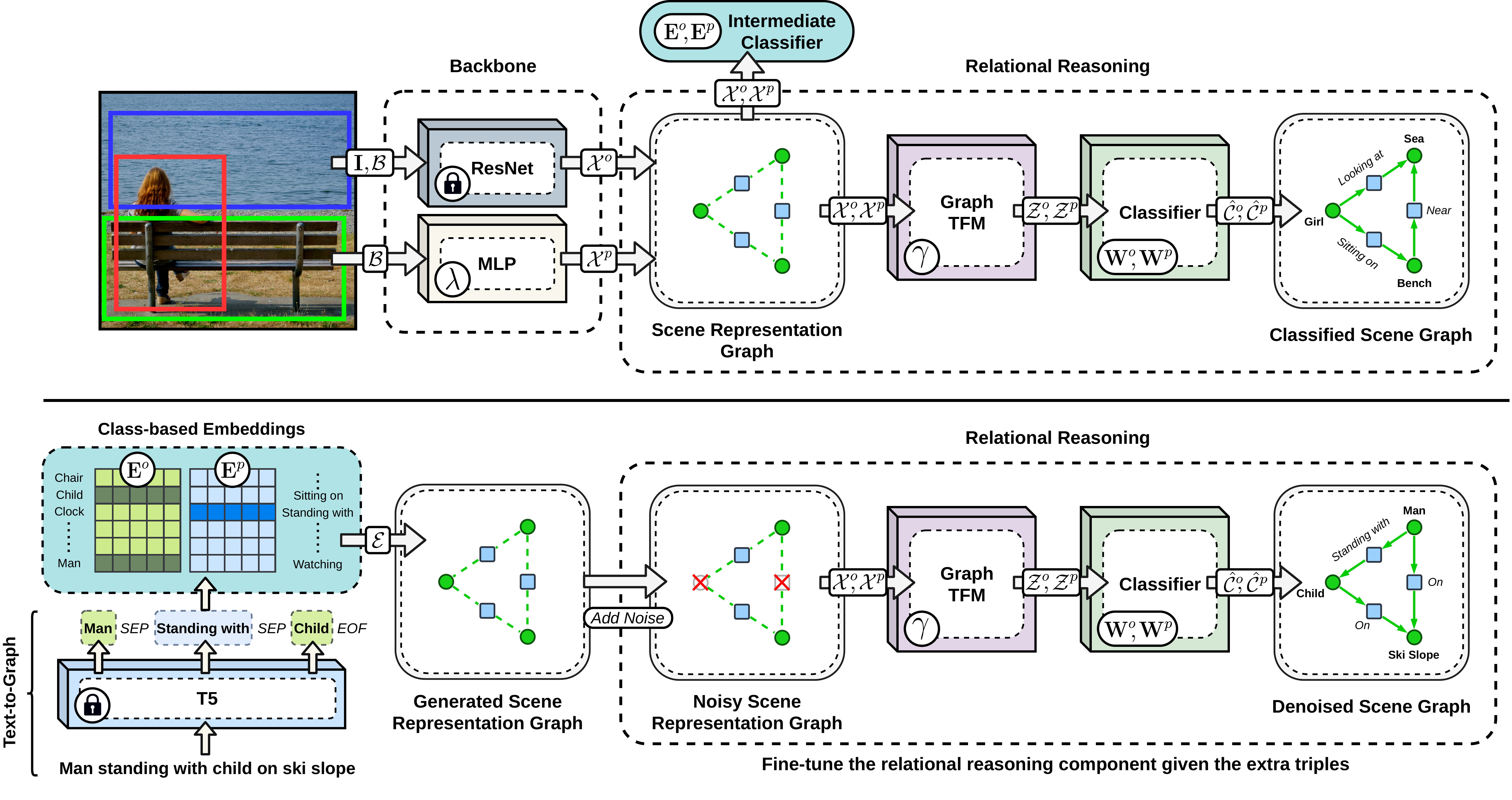}
	\end{center}
	\caption{ Top: we initially train a scene graph classification pipeline from images and their corresponding SGs.
	Bottom: we then use a text-to-graph module to extract structured knowledge from unstructured texts. The extracted graph is embedded by image-grounded vectors, masked, and then fed to the relational reasoning module to predict the missing relations and thus, encourage the network to learn the new relations from texts. The \textit{lock} sign indicates pre-trained and frozen parts of the network.}\label{arch}
\end{figure*}

We consider a classification pipeline with two major parts: a feature extraction \textit{backbone}, and a \textit{relational reasoning} component (Figure \ref{arch}). The backbone is typically a convolutional neural network (CNN) that detects objects and extracts an image-based representation for each. On the other hand, the relational reasoning component can be a variant of a recurrent neural network~\citep{xu2017scene, zellers2018neural} or graph convolutional networks~\citep{yang2018graph, sharifzadeh2021classification}. This component operates on an object level  by taking the latent representations of all the objects in the image and propagating them in the graph.
 
Note that, unlike the feature extraction backbone that requires images as input, the relational reasoning component operates on graphs with the nodes representing objects and the edges representing relations. The distinction between the input to the backbone (images) and  the relational reasoning component (graphs) is often overlooked. Instead, the scene graph classification pipeline is treated as a network that takes only images as inputs. However, one can also train or fine-tune the relational reasoning component directly  by injecting it with relational knowledge. For example, Knowledge Graphs (KGs) contain curated facts that indicate the relations between a \texttt{head} object and a \texttt{tail} object in the form of \texttt{(head, predicate, tail)} e.g., \texttt{(Person, Rides, Horse)}. The facts in KGs are represented by symbols whereas the inputs to the relational reasoning component are image-based embeddings. In this work, we map the triples to image-grounded embeddings as if they are coming from an image. We then use these embeddings to fine-tune the relational reasoning component through a denoising graph autoencoder scheme.

Note that the factual knowledge is not always available in a well-structured form, specially in domains where the knowledge is not stored in the machine-accessible form of KGs. In fact, most of the collective human knowledge is only available in the unstructured form of texts and documents. Exploiting this form of knowledge, in addition to structured knowledge, can be significantly beneficial. To this end, we employ a transformer-based model to generate structured graphs from textual input and utilize them to improve the relational reasoning module. 

In summary, we propose \textit{Texema}, a scene graph classification pipeline that can be trained from the large corpora of unstructured knowledge. We evaluate our approach on the Visual Genome dataset. In particular, we show that we can fine-tune the reasoning component using textual scene descriptions instead of thousands of images. As a result, when using as little as {\raise.12ex\hbox{$\scriptstyle\sim$}}500 images (1\% of the VG training data), we can achieve {\raise.12ex\hbox{$\scriptstyle\sim$}}3x more accurate results in object classification, {\raise.12ex\hbox{$\scriptstyle\sim$}}8x in scene graph classification and {\raise.12ex\hbox{$\scriptstyle\sim$}}1.5x in predicate classification compared to the supervised baselines. Additionally, in our ablation studies, we evaluate the performance of using different rule-based, LSTM-based, and transformed-based text-to-graph models.

%Related Works
\section{Related Works}

\textbf{Scene Graph Classification:} There is an extensive body of work on visual reasoning in general that includes different forms of reasoning~\citep{wu2014hierarchical,deng2014large,hu2016learning,hu2017labelbank,santoro2017simple, zellers2019recognition}. Here, we mainly review the works that are focused on scene graph classification. Visual Relation Detection (VRD)~\citep{lu2016visual} and the Visual Genome~\citep{krishna2017visual} are the main datasets for this task. While the original papers on VRD and VG provide the baselines for scene graph classification by treating objects independently, several follow-up works contextualize the entities before classification. Iterative Message Passing (IMP)~\citep{xu2017scene}, Neural Motifs~\citep{zellers2018neural} (NM), Graph R-CNN~\citep{yang2018graph}, and Schemata~\citep{sharifzadeh2021classification} proposed to propagate the image context using basic RNNs, LSTMs, graph convolutions, and graph transformers respectively. On the other hand, authors of VTransE~\citep{zhang2017visual} proposed to capture relations by applying TransE~\citep{bordes2013translating}, a knowledge graph embedding model, on the visual embeddings, ~\citet{tang2019learning} exploited dynamic tree structures to place the object in an image into a visual context. ~\citet{chen2019counterfactual} proposed a multi-agent policy gradient method that frames objects into cooperative agents and then directly maximizes a graph-level metric as the reward. In tangent to those works, ~\citet{9412945} proposed to enrich the input domain in scene graph classification by employing the predicted pseudo depth maps of VG images that were released as an extension called \textit{VG-Depth}.
\\
\textbf{Commonsense in Scene Understanding: }
Several recent works have proposed to employ external or internal sources of knowledge to improve visual understanding~\citep{wang2018zero, jiang2018hybrid, singh2018dock, kato2018compositional}. In the scene graph classification domain, some of the works have proposed to correct the SG prediction errors by merely comparing them to the co-occurrence statistics of internal triples as a form of commonsense knowledge~\citep{chen2019knowledge,chen2019learning, zellers2018neural}. Earlier, ~\citet{baier2017improving,baier2018improving} proposed the first scene graph classification model that employed prior knowledge in the form of Knowledge Graph Embeddings (KGEs) that generalize beyond the given co-occurrence statistics. ~\citet{zareian2020bridging,zareian2020learning} followed this approach by extending it to models that are based on graph convolutional networks. More recently, ~\citet{sharifzadeh2021classification} proposed Schemata as a generalized form of a KGE model that is learned directly from the images rather than triples. In general, scene graph classification methods are closely related to the KGE models. Therefore, we refer the interested readers to~\citep{nickel2016review,ali2020bringing, ali2020pykeen} for a review and large-scale study on the KG models, and  to~\citep{tresp2019model,tresp2020tensor} for an extensive investigation of the connection between perception, KG models, and cognition.

Nevertheless, to the best of our knowledge, the described methods have employed curated knowledge in the form of triples, and none of them have directly exploited the textual knowledge. In this direction, the closest work to ours is by~\citet{yu2017visual}, proposing to distill the external language knowledge using a teacher-student model. However, this work does not include a relational reasoning component and only refines the final predictions. Also, as shown in the experiments, our knowledge extraction module performs two times better than the SG Parser used in that work.
\\
\textbf{Knowledge Extraction from Text: }
Knowledge extraction from text has been studied for a long time \citep{chinchor-1991-muc-3}.
Previous work ranges from pattern-based approaches \citep{hearst-1992-automatic} to supervised neural approaches with specialized architectures \citep{gupta-etal-2019-neural,yaghoobzadeh-etal-2017-noise}.
Recently, \citet{schmitt2020unsupervised} successfully applied a general sequence-to-sequence architecture to graph$\leftrightarrow$text conversion.
With the recent rise of transfer learning in NLP,
an increasing number of approaches are based on large language models, pre-trained in a self-supervised manner on massive amounts of texts \citep{devlin-etal-2019-bert}.
Inspired from previous work that explores transfer learning for graph-to-text conversion \citep{ribeiro-etal-2020-investigating},
we base our text-to-graph model on a pre-trained T5 model \citep{raffel2019exploring}.

%Methods
\begin{figure}[t]
	\begin{center}
	\includegraphics[width=0.35\textwidth]{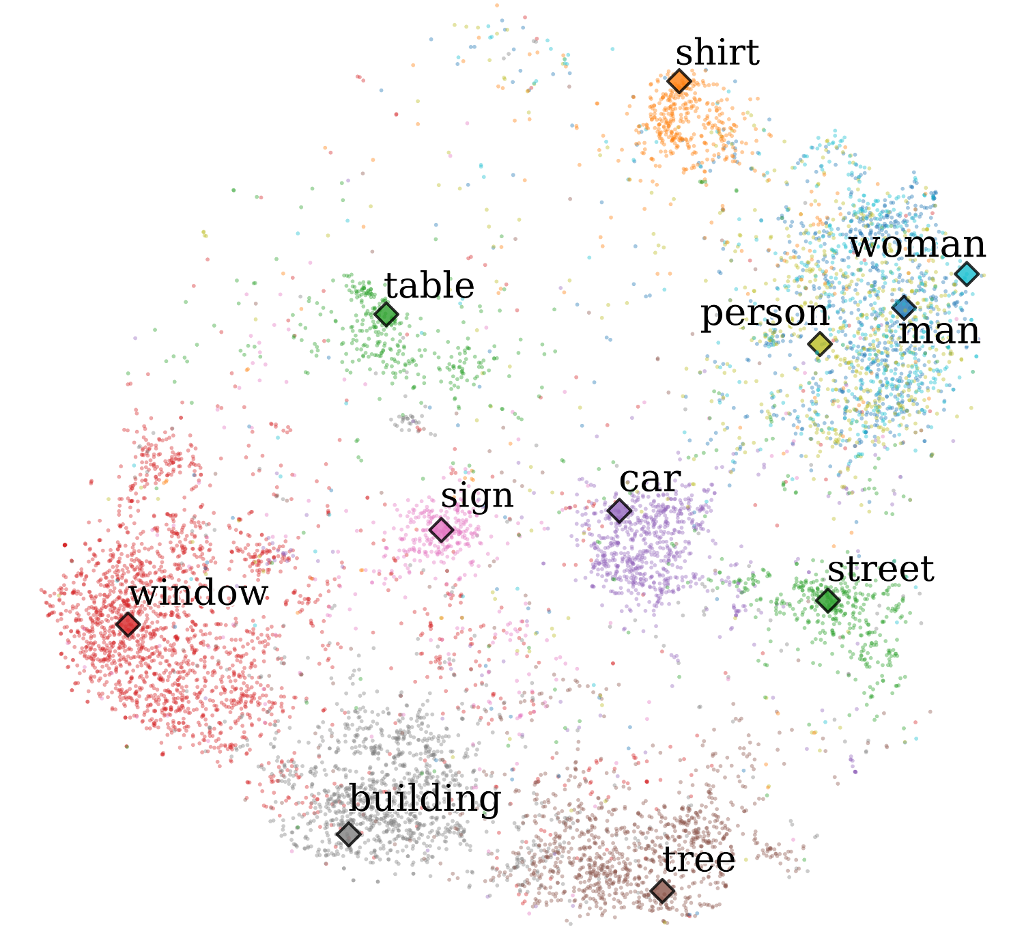}
	\end{center}
    \caption{The t-SNE representation of the $\mathbf{e}_i$s (diamonds) and image-based representations $\mathcal{X}$s (dots) where each color represents the ground-truth class of the dot.}
\label{fig_tsne}
\end{figure}

% \begin{table}
% 	\centering
%     \small
% 	\begin{tabularx}{1.0\linewidth}{lX}
% 		\toprule
% 		\textbf{Input} &  \textbf{man standing with child on ski slope} \\
% 		\midrule
% 		Reference & \mbox{(child, on, ski slope)} \\
% 		Graph (RG) & \mbox{(man, standing with, child)} \mbox{(man, on, ski slope)} \\
% 		\midrule
% 		\rttg{} & \colorbox{red}{(man, standing, child)} \\
% 		\midrule
% 		Stanford Scene & \colorbox{red}{(standing, with, child)}, \\
% 		Graph Parser   & \colorbox{red}{(standing, on, slope)} \\
% 		\midrule
% 		CopyNet (1\%){} & \colorbox{green}{(man, standing with, child)} \\
% 		\midrule
% 		T5 (1\%){} & \colorbox{green}{(man, standing with, child)} \\
% 		\midrule
% 		CopyNet (10\%){} & \colorbox{green}{(man, standing with, child)}\\
% 		& \colorbox{red}{(child, on, slope)} \\
% 		\midrule
% 		T5 (10\%){} & \colorbox{green}{(man, standing with, child)} \\
% 		& \colorbox{green}{(child, on, ski slope)}\\
% 		\bottomrule
		
% 	\end{tabularx}
	
%     \caption{Example fact extractions and evaluation wrt.\ reference graph (RG).
% 		Green: correct ($\in$ RG). Red: incorrect ($\notin$ RG). The results are computed using our re-implementation of the related works and evaluated on the Visual Genome splits.}
% 		\label{tab:ie-qual}
% \end{table}

\begin{table}
	\centering
    \small
\setlength{\tabcolsep}{3pt}
	\begin{tabularx}{1.0\linewidth}{lX}
		\toprule
		\textbf{Input} &  \textbf{man standing with child on ski slope}  \\
		\midrule
		Reference & \mbox{(child, on, ski slope)} \mbox{(man, on, ski slope)}\\
		Graph (RG) & \mbox{(man, standing with, child)} \\
		\midrule
		\rttg{} & \colorbox{red}{(man, standing, child)} \\
		\midrule
		SSGP& \colorbox{red}{(standing, with, child)}\colorbox{red}{(standing, on, slope)}\\
		%Graph Parser   & \colorbox{red}{(standing, on, slope)} \\
		\midrule
		CopyNet (1\%){} & \colorbox{green}{(man, standing with, child)} \\
		\midrule
		T5 (1\%){} & \colorbox{green}{(man, standing with, child)} \\
		\midrule
		CopyNet (10\%){} & \colorbox{green}{(man, standing with, child)}\colorbox{red}{(child, on, slope)}\\
		%& \colorbox{red}{(child, on, slope)} \\
		\midrule
		T5 (10\%){} & \colorbox{green}{(man, standing with, child)} \\
		& \colorbox{green}{(child, on, ski slope)}\\
		\bottomrule
		
	\end{tabularx}
	
    \caption{An example of extracted triples from a given text input in VG, using different methods. Green: correct ($\in$ RG). Red: incorrect ($\notin$ RG). The results are computed using the respective official code bases of the related works.}
		\label{tab:ie-qual}
\end{table}

\begin{algorithm}[t]
\caption{Classify objects/predicates from images}
\label{alg:1}
    {\small
\begin{enumerate}[leftmargin=*]
    \item \textbf{Extract image features (Backbone):}

        \begin{description}
            \item \textbf{Input:} Images and object bounding boxes ($\mathbf{I},\mathcal{B}:\lbrace \mathbf{b_i} \rbrace_{i=1}^{n}$).
            \item \textbf{Output:} Object embeddings $\mathcal{X}^o:\lbrace \mathbf{x}^o_i \rbrace_{i=1}^{n}$ and predicate embeddings $\mathcal{X}^p:\lbrace \mathbf{x}^p_i \rbrace_{i=1}^{m}$.
            \item \textbf{Trainable params:} $\lambda$.
        \end{description}
        \begin{description}
            \item $\mathcal{X}^o = ResNet50(\mathbf{I}, \mathcal{B})$ 
            \item $\mathcal{X}^p=\lbrace MLP_{\lambda}(t(\mathbf{b}_i,  \mathbf{b}_j)) \mid \forall \mathbf{b}_i, \mathbf{b}_j \in \mathcal{B} \rbrace $
        \end{description}
    
    \item \textbf{Contextualize and Classify (Relational Reasoning):}
 
        \begin{description}
            \item \textbf{Input:} Object embeddings $\mathcal{X}^o:\lbrace \mathbf{x}^o_i \rbrace_{i=1}^{n}$, Predicate embeddings $\mathcal{X}^p:\lbrace \mathbf{x}^p_i \rbrace_{i=1}^{m}$ and ground truth classes $\mathcal{C}^o$ and $\mathcal{C}^p$.
            \item \textbf{Output:} Predicted object class distribution $\hat{\mathcal{C}}^o:\lbrace \hat{\mathbf{c}}^o_i \rbrace_{i=1}^{n}$ and predicted predicate class distribution $\hat{\mathcal{C}}^p:\lbrace \hat{\mathbf{c}}^p_i \rbrace_{i=1}^{m}$.
            \item \textbf{Trainable params:} $\gamma$, $\mathbf{W}^o, \mathbf{W}^p$.
        \end{description}    
        \begin{description}
            \item $\mathcal{Z}^o, \mathcal{Z}^p = GraphTransformer_{\gamma}(\mathcal{X}^o, \mathcal{X}^p)$ 
            \item $\hat{\mathcal{C}}^o = \lbrace \textrm{softmax}({\mathbf{W}^o} \cdot \mathbf{z}^o) \mid \forall \mathbf{z}^o \in \mathcal{Z}^o\rbrace$
            \item $\hat{\mathcal{C}}^p = \lbrace \textrm{softmax}(\mathbf{W}^p \cdot \mathbf{z}^p) \mid \forall \mathbf{z}^p \in \mathcal{Z}^p\rbrace$
        \end{description}    
    
    \item \textbf{Apply Loss (Cross-Entropy):}

        \begin{description}

            \item $l_o = - \frac{1}{n}  \sum_{i=1}^{n}   \sum_{j=1}^{\|\mathbf{c}^o_i\|}    \mathbf{c}^o_{i,j}.log({\hat{\mathbf{c}}}^o_{i,j})$ 

            \item $l_p = - \frac{1}{m}  \sum_{i=1}^{m}   \sum_{j=1}^{\|\mathbf{c}^p_i\|}    \mathbf{c}^p_{i,j}.log({\hat{\mathbf{c}}}^p_{i,j})$
        \end{description}

\end{enumerate}
}
\end{algorithm}
\begin{algorithm}[t]
\caption{Fine-tune the relational reasoning component from textual triples using a denoising auto-encoder paradigm}
\label{alg:2}
\begin{enumerate}[leftmargin=*]
    {\small
    \item \textbf{Learn image-grounded representations $\mathbf{E}$ for each symbol through classification (without Graph Transformer):}
    \begin{description}
        \item \textbf{Input:} Object embeddings $\mathcal{X}^o:\lbrace \mathbf{x}^o_i \rbrace_{i=1}^{n}$, predicate embeddings $\mathcal{X}^p:\lbrace \mathbf{x}^p_i \rbrace_{i=1}^{m}$ and their corresponding ground truth classes $\mathcal{C}^o$ and $\mathcal{C}^p$.
        \item \textbf{Output:} Predicted object class distribution $\hat{\mathcal{C}}^o:\lbrace \hat{\mathbf{c}}^o_i \rbrace_{i=1}^{n}$ and predicted predicate class distribution $\hat{\mathcal{C}}^p:\lbrace \hat{\mathbf{c}}^p_i \rbrace_{i=1}^{m}$.
        \item \textbf{Trainable params:} $\mathbf{E}^o, \mathbf{E}^p$.
    \end{description}    
    \begin{description}
        \item $\hat{\mathcal{C}}^o = \lbrace \textrm{softmax}(\mathbf{E}^o \cdot \mathbf{x}^o) \mid \forall \mathbf{x}^o \in \mathcal{X}^o\rbrace$
        \item $\hat{\mathcal{C}}^p = \lbrace \textrm{softmax}(\mathbf{E}^p \cdot \mathbf{x}^p) \mid \forall \mathbf{x}^p \in \mathcal{X}^p\rbrace$
    \end{description}    

\item \textbf{Apply Loss (Cross Entropy):}

    \begin{description}

        \item $l_o = - \frac{1}{n}  \sum_{i=1}^{n}   \sum_{j=1}^{\|\mathbf{c}^o_i\|}    \mathbf{c}^o_{i,j}.log({\hat{\mathbf{c}}}^o_{i,j})$ 

        \item $l_p = - \frac{1}{m}  \sum_{i=1}^{m}   \sum_{j=1}^{\|\mathbf{c}^p_i\|}    \mathbf{c}^p_{i,j}.log({\hat{\mathbf{c}}}^p_{i,j})$
    \end{description}

    \item \textbf{Fine-tune the relational reasoning component given the extra triples (Denoising Graph Autoencoder):}

        \begin{description}
            \item \textbf{Input:} Symbolic triples $\mathcal{S}:\lbrace (h_i, p_i, t_i) \rbrace_{i=1}^{k}$ and canonical object/predicate representations $\mathbf{E}^o$/$\mathbf{E}^p$.
            \item \textbf{Output:} Embedded representations $\mathcal{E} : \lbrace (\mathbf{e}^h_i, \mathbf{e}^p_i, \mathbf{e}^t_i) \rbrace_{i=1}^{k} $.
            \item \textbf{Trainable params:} $\gamma, \mathbf{W}^o, \mathbf{W}^p$.
        \end{description}    
        
        \begin{itemize}
        \item
        \begin{description}%[leftmargin=*]
                    \item Build $\mathcal{E} : \lbrace (\mathbf{e}^h_i, \mathbf{e}^p_i, \mathbf{e}^t_i) \rbrace_{i=1}^{k} $ where for each $(h_i, p_i, t_i)$:% 
                 \item$\mathbf{e}_i^h = onehot(h_i)\cdot \mathbf{E}^o$
                 \item$\mathbf{e}_i^p = onehot(p_i)\cdot \mathbf{E}^p$
                \item$\mathbf{e}_i^{t} = onehot(t_i)\cdot \mathbf{E}^o$
            \end{description}
 
            \item Randomly set $20\%$ of the nodes and edges in $\mathcal{E}$ to zero.
            \item Set $\mathcal{X}^o = \mathcal{E}^h \cup \mathcal{E}^t$ and $\mathcal{X}^p = \mathcal{E}^p$ and run Algorithm 1.2 to fine-tune $\gamma, \mathbf{W}^o, \mathbf{W}^p$, with $\mathcal{E}^h$, $\mathcal{E}^t$ and $\mathcal{E}^p$ as the set of all heads, tails, and predicates in $\mathcal{E}$.
        \end{itemize}    
    }

\end{enumerate}
\end{algorithm}

\section{Methods}
In this section, we first describe the backbone and relational reasoning components. We then describe our approach for fine-tuning the network from texts.
We have three possible forms of data:  Images (\textbf{IM}), Scene Graphs (\textbf{SG}) and Textual Scene Descriptions (\textbf{TXT}). We consider having two sets of data: one is the \textit{parallel} set, which is the set of IM with their corresponding SG and TXT, and another is the \textit{text} set which is a set of additional TXT that come without any images or scene graphs. These two sets have no elements in common.

We initially train our backbone and relational reasoning component from IM and SG, and our text-to-graph model from the TXT and SG in the parallel set. We then show that we can fine-tune the pipeline using the text set and without using any additional images.

\subsection{Backbone (Algorithm 1.1)}
The feature-extraction backbone is a convolutional neural network (ResNet-50) that has been pre-trained in a self-supervised manner~\citep{grill2020bootstrap} from unlabeled images of ImageNet~\citep{deng2009imagenet} and Visual Genome~\citep{krishna2017visual}. Given an image $\mathbf{I}$ with several objects in bounding boxes $\mathcal{B} = \{\mathbf{b}_i\}_{i=1}^n$, $\mathbf{b}_i=[b^x_i,b^y_i,b^w_i,b^h_i]$, we apply the ResNet-50 to extract pooled object features $\mathcal{X}^o = \{\mathbf{x}_i^o\}_{i=1}^n$, $\mathbf{x}^o_i \in \mathbb{R}^d$. Here $[b^x_i,b^y_i]$ are the coordinates of $\mathbf{b}_i$ and $[b^w_i,b^h_i]$ are its width and height, and $d$ are the vector dimensions. Following ~\citep{zellers2018neural}, we define $\mathcal{X}^p= \{\mathbf{x}_i^p\}_{i=1}^m$, $\mathbf{x}^p_i \in \mathbb{R}^d$ as the relational features between each pair of objects. Each $\mathbf{x}^p_i$ is initialized by applying a two layered fully connected network on the relational position vector $\mathbf{t}$ between a head $i$ and a tail $j$ where $\mathbf{t} = [t_x, t_y, t_{w}, t_{h}]$, 
$
t_x = (b^x_{i} - b^x_{j})/{b^w_i}_{j}, t_y = (b^y_{i} - b^y_{j})/b^h_{j}, t_{w} = \log(b^w_{i}/b^w_{j}), t_{h} = \log(b^h_{i}/b^h_{j})
$. The implementation and pre-training details of the layers are provided in the Evaluation. $\mathcal{X}^o$ and $\mathcal{X}^p$ form a structured presentation of the objects and predicates in the image also known as \textbf{Scene Representation Graph} (SRG)~\citep{sharifzadeh2021classification}. SRG is a fully connected graph with each node representing either an object or a predicate, where each object node is a direct neighbor to predicate nodes and each predicate node is a direct neighbor with its head and tail object nodes.

\subsection{Relational Reasoning (Algorithm 1.2)}
The relational reasoning component updates the initial SRG representations through Graph Transformer layers~\citep{koncel2019text}. 
The outputs of these layers are $\mathcal{Z}^o = \{\mathbf{z}_i^o\}_{i=1}^n$, $\mathbf{z}^o_i \in \mathbb{R}^d$ and $\mathcal{Z}^p= \{\mathbf{z}_i^p\}_{i=1}^m$, $\mathbf{z}^p_i \in \mathbb{R}^d$ with equal dimensions as $\mathcal{X}$s. From here on, we drop the superscripts of $o$ and $p$ for brevity. We apply a linear classification layer $\mathbf{W}$ to classify the contextualized representations $\mathcal{Z}$ such that $\hat{\mathbf{c}} = \textrm{softmax}(\mathbf{W} \cdot \mathbf{z}_i)$, with cross-entropy as the loss function. 

\subsection{Fine-tuning from Texts (Algorithm 2)} Let us assume that we have already trained the backbone and relational reasoning components from IM and SG in the \textit{parallel} set. Now, we want to fine-tune the weights in the relational reasoning component given the additional \textit{text} set. The relational reasoning component takes graphs as input, therefore, we first need to convert TXT to SG:
\paragraph{Text-to-graph:} This model is trained from the SG and TXT in the parallel set, and then used to generate SG from the text set. Let us consider an unstructured text such as ``man standing with child on ski slope'' (Table \ref{tab:ie-qual} - Input). A structured form of this sentence is a graph with unique nodes and edges for each entity or predicate. For example, the reference graph for this sentence contains the triples \texttt{(child, on, ski slope)}, \texttt{(man, standing with, child)} and \texttt{(man, on, ski slope)} (Table \ref{tab:ie-qual} - RG). 

In order to learn this mapping, we employ a transformer-based~\citep{vaswani2017attention} sequence-to-sequence T5\textsubscript{small} model~\citep{raffel2019exploring} and adapt it for the task of extracting graphs from texts. T5 consists of an encoder with several layers of self-attention (like BERT, \citealp{devlin-etal-2019-bert}) and a decoder with autoregressive self-attention (like GPT-3, \citealp{brown20}).
In order to use a T5 model with graphs, we need to represent the graphs as a sequence. To this end, we serialize the graphs by writing out their facts separated by end-of-fact symbols (\texttt{EOF}), and separate the elements of each fact with \texttt{SEP} symbols~\citep{schmitt2020unsupervised}, e.g. \textit{``child SEP on SEP ski slope EOF''} (Fig. \ref{arch}). To adapt the multi-task setting from T5's pretraining, we use the task prefix ``make graph: '' to mark our text-to-graph task.
Table \ref{tab:ie-qual} shows an example text and the extracted graphs using T5 and other previous methods (see Evaluation for details).

\paragraph{Map to embeddings:}
Note that the predicted graphs are a sequence of symbols for heads, predicates, and tails where each symbol represents a class $c \in \mathcal{C}$. However, the inputs to the relational reasoning component are image-based vectors $\mathcal{X}$. Thus, before feeding the symbols to the relational reasoning component, we need to map them to a corresponding embedding from the space of $\mathcal{X}$ as if we are feeding it with image-based embeddings. In order to do that, we train a mapping from symbols to $\mathcal{X}$s using the IM and SG of the parallel set. This is simply done by training a linear classification layer $\mathbf{E}$ given $\mathcal{X}$s from the parallel set (Algorithm 2.1). Unlike the classification layer in Algorithm 1, here we classify $\mathcal{X}$s instead of $\mathcal{Z}$s and the goal is \textit{not} to use the classification output but to train image-grounded, canonical representations for each class: each row $\mathbf{e}_i$ in the classification layer becomes a cluster center for $\mathcal{X}$s from class $i$ (Figure \ref{fig_tsne}). Therefore, instead of the extracted symbolic $c_i$ from the text set, we can feed its canonical image-grounded representation $\mathbf{e}_i$ to the graph transformer (Algorithm 2.3). 
\paragraph{Denoising Graph Autoencoder:} To fine-tune the relational reasoning given this data, we treat the relational reasoning component as a denoising autoencoder where the input is an incomplete (noisy) graph that comes from the text and the output is the denoised graph. If we do not apply a denoising autoencoder paradigm, the function will collapse to an identity map. We create the noisy graph by randomly setting some of the input nodes and edges to zero during the training (Algorithm 2.3). The goal is to encourage the graph transformer to predict the missing links and therefore, learn the relational structure.

%Evaluation
% Text2Graph Table (full row)
{\renewcommand{\arraystretch}{1.35}
\begin{table*}[t] 
	\centering
    \small
	\scalebox{1.0}{
		\begin{tabular}{c|cc|cc|cc}
			%\toprule
			\hline
			\multirow{2}[1]{*}{Method}& \multicolumn{2}{c|}{Precision} & \multicolumn{2}{c|}{Recall} & \multicolumn{2}{c}{F1} \\
			%\cmidrule(lr){2-3}\cmidrule(lr){4-5}\cmidrule(lr){6-7}
			&$1\%$ & $10\%$ &$1\%$ & $10\%$ &$1\%$ & $10\%$ \\
			\hline
			\hline
			\rttg{} & $1.92\pm0.00$ & $1.86\pm 0.01$ & $1.87\pm0.00$ & $1.81\pm 0.01$ & $1.89\pm0.00$ & $1.84\pm 0.01$  \\
			SSGP  & $14.86\pm0.01$ & $14.52\pm0.02$ & $18.47\pm0.01$ & $18.05\pm0.02$
			& $16.47\pm0.01$ & $16.09\pm0.02$ \\
			CopyNet & $29.20\pm0.13$ & $30.77\pm0.49$ & $27.19\pm0.28$  & $29.79\pm0.29$
			& $28.16\pm0.21$ & $30.27\pm0.34$ \\
			\textbf{T5} & $\textbf{33.37}\pm\textbf{0.11}$ & $\textbf{33.81}\pm\textbf{0.08}$ & $\textbf{31.06}\pm\textbf{0.18}$ & $\textbf{32.45}\pm\textbf{0.33}$ & $\textbf{32.17}\pm\textbf{0.13}$ & $\textbf{33.12}\pm\textbf{0.16}$  \\
			\hline
		\end{tabular}
	}
	\caption{The mean and standard deviation of Precision, Recall, and F1 scores of the predicted facts from the texts on four random splits. The results are computed using the respective official code bases of the related works and evaluated on VG.}
	\label{tab:ie-quant}
\end{table*}}
\section{Evaluation}
We first compare the performance of different rule-based and embedding-based text-to-graphs models on our data. We then evaluate the performance of our entire pipeline in classifying objects and relations in images. In particular, we show that the extracted knowledge from the texts can largely substitute annotated images as well as ground-truth graphs.% We ran all experiments on NVIDIA GeForce RTX 2080 Ti GPUs with 11 GB VRAM.
\paragraph{Dataset}
We use the sanitized version~\citep{xu2017scene} of Visual Genome (VG) dataset~\citep{krishna2017visual} including images and their annotations, i.e., bounding boxes, scene graphs, and scene descriptions. Our goal is to design an experiment that evaluates whether we can substitute annotated images with textual scene descriptions. Therefore, instead of using external textual datasets with unbounded information, we use Visual Genome itself by dividing it into different splits of \textit{parallel} (with IM, SG and TXT) and \textit{text} data (with only TXT). To this end, we assume only a random proportion (1\% or 10\%) of training images are annotated (parallel set containing IM with corresponding SG and TXT). We consider the remaining data (99\% or 90\%) as our text set and discard their IM and SG. We aim to see whether employing TXT from the text set, can substitute the discarded IM and SG from this set. We use four different random splits~\citep{sharifzadeh2021classification} to avoid a sampling bias. For more detail on the datasets refer to the supplementary materials.

Note that the scene graphs and the scene descriptions from the VG are collected separately and by crowd-sourcing. Therefore, even though the graphs and the scene descriptions refer to the same image region, they are disjoint and contain complementary knowledge.

\subsection{Graphs from Texts}
The goal of this experiment is to study the effectiveness of the text-to-graph model. We fine-tune the pre-trained T5 model on parallel TXT and SG, and apply it on the text set to predict their corresponding SG.
We also implement the following rule-based and embedding-based baselines to compare their performance using our splits:
(1) \rttg{} is a simple rule-based system introduced by \citet{schmitt2020unsupervised} for general knowledge graph generation from text.
(2) The Stanford Scene Graph Parser (SSGP)~\citep{schuster2015generating} is another rule-based approach that is more adapted to the scene graph domain. Even though this approach was not specifically designed to match the scene graphs from the Visual Genome dataset, it was still engineered to cover typical idiosyncrasies of textual image descriptions and corresponding scene graphs.
(3) CopyNet \citep{gu-etal-2016-incorporating} is an LSTM sequence-to-sequence model with a dedicated copy mechanism, which allows copying text elements directly into the graph output sequence.
It was used for unsupervised text-to-graph generation by \citet{schmitt2020unsupervised}.
However, we train it on the supervised data of our parallel sets.
We use a vocabulary of around 70k tokens extracted from the VG-graph-text benchmark \citep{schmitt2020unsupervised}
and, otherwise, also adopt the hyperparameters from \citep{schmitt2020unsupervised}. Table \ref{tab:ie-qual} shows sample predictions from these models. Table \ref{tab:ie-quant} compares their precision, recall, and F1 measures. T5 outperforms other models by a large margin.

% ablation table
{\renewcommand{\arraystretch}{1.35}
	\begin{table*}[!t]
		\centering
		\small
		\scalebox{1.0}{
			\begin{tabular}{c|c|cc|cc}
				\hline
				\multirow{2}{*}{} &\centering \multirow{2}{*}{Method} &
				\multicolumn{2}{c|}{R@50} & \multicolumn{2}{c}{R@100}  \\
				& \multirow{2}{*}  & $1{\%}$ & $10{\%}$ & $1{\%}$ & $10{\%}$\\
				\hline
				\hline
				\multirow{5}{*}{\rotatebox[origin=c]{0}{SGCls}} 
                & \centering \rttg{}    & $10.90\pm0.12$ & $24.96\pm0.15$ & $11.80\pm0.11$& $26.09\pm0.15$ \\
				& \centering SSGP       & $14.35\pm0.15$ & $26.11\pm0.19$ & $15.14\pm0.17$& $27.12\pm0.22$ \\
				& \centering CopyNet    & $14.46\pm0.31$ & $26.05\pm0.29$ & $15.19\pm0.24$& $27.08\pm0.26$ \\
	            & \centering \textbf{TXM - T5}  & $\mathbf{14.53 \pm 0.34}$ & $\mathbf{26.16 \pm 0.32}$ & $\mathbf{15.28 \pm 0.38}$& $\mathbf{27.22 \pm 0.28}$\\
				\cdashline{2-6}
				& \centering GT &  ${14.72\pm0.38}$ & ${26.33\pm0.45}$ & ${15.36\pm0.38}$ & ${27.37\pm0.47}$\\
				% & \centering FSPB & $\mathbf{14.53 \pm 0.34}$ & $\mathbf{26.16 \pm 0.32}$ & $\mathbf{15.28 \pm 0.38}$& $\mathbf{27.22 \pm 0.28}$\\
				\hline

				\multirow{5}{*}{\rotatebox[origin=c]{0}{PredCls}} 
                & \centering \rttg{}    & $23.34\pm0.10$ & $49.99\pm0.12$ & $26.83\pm0.15$& $54.40\pm0.12$ \\
				& \centering SSGP       & $54.65\pm0.14$ & $55.65\pm0.15$ & $59.33\pm0.18$& $59.67\pm0.20$ \\
				& \centering CopyNet    & $56.24\pm0.31$ & $59.27\pm0.28$ & $60.35\pm0.20$& $63.28\pm0.25$ \\
				& \centering \textbf{TXM - T5} & $\mathbf{58.64 \pm 0.34}$ & $\mathbf{59.31 \pm 0.30}$ & $\mathbf{63.07 \pm 0.37}$& $\mathbf{63.32 \pm 0.24}$\\
				\cdashline{2-6}
				& \centering GT & ${62.02\pm0.10}$ & ${61.71\pm0.19}$ & ${65.68\pm0.12}$ & ${65.42\pm0.19}$\\
				% & \centering FSPB & $\mathbf{14.53 \pm 0.34}$ & $\mathbf{26.16 \pm 0.32}$ & $\mathbf{15.28 \pm 0.38}$& $\mathbf{27.22 \pm 0.28}$\\
				\hline
				
		\end{tabular}}
		\caption{SGCls and PredCls results using different text-to-graph modules. We have substituted the missing 99\% and 90\% of annotated images with the textual knowledge extracted from their scene descriptions.}
		\label{full_results}
	\end{table*}
}

\begin{figure}[t]
	\begin{center}
		\includegraphics[width=0.45\textwidth]{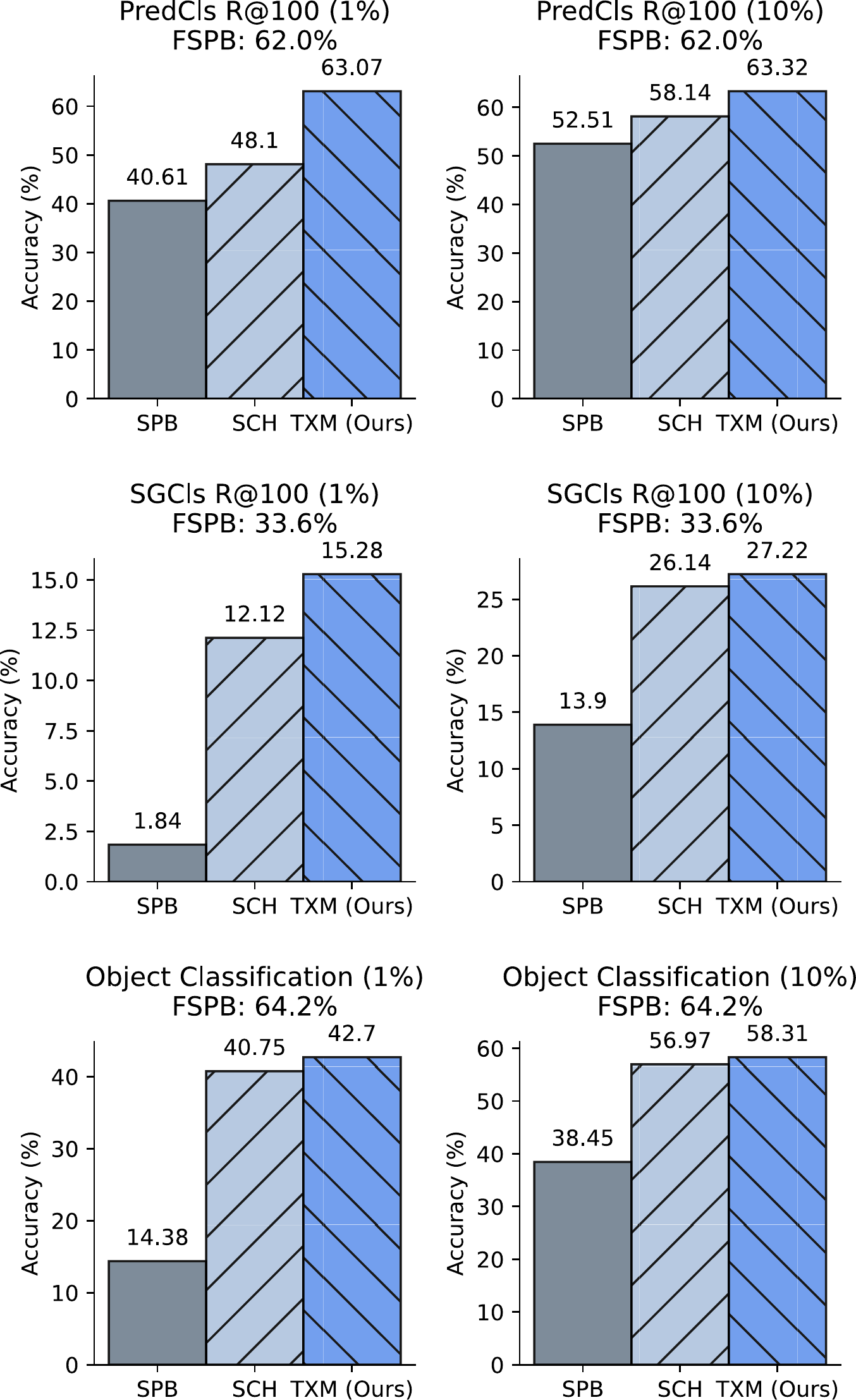}

	\end{center}
	\caption{Fine-tuning with the textual knowledge (TXM) significantly improves the results in all settings of PredCls (top), SGCls (middle), and object classification (bottom).}	
	\label{fig_golden_obj}
\end{figure}

\subsection{Graphs from Images}
The goal of this experiment is to evaluate the performance of scene graph classification after fine-tuning the pipeline using textual knowledge only. We evaluate our models for object classification, predicate classification (PredCls - predicting predicate labels given a ground truth set of object boxes and object labels) and scene graph classification (SGCls - predicting object and predicate labels, given the set of object boxes) on the test sets. Since the focus of our study is not improving object detection, we skip scene graph detection. Our ablation study concerns the following configurations:

\begin{itemize}[leftmargin=*]

\item\textbf{SPB}: In this setting, both the backbone and the relational reasoning component are trained by \textit{supervised learning} on the IM and SGs (1\% or 10\%) from the parallel set.

\item\textbf{SCH}: Here, the backbone is trained by \textit{self-supervised learning} on all VG images (without labels), and the relational reasoning component is trained on the IM and SGs (1\% or 10\%) from the parallel set.

\item\textbf{TXM}: Here, the backbone is trained by \textit{self-supervised learning} on all VG images (without labels), and the relational reasoning component is trained on the IM and SGs (1\% or 10\%) from the parallel set and fine-tuned from the SGs predicted from the text set (99\% or 90\%) using the text-to-graph module. 

\item\textbf{GT}: Here, the backbone is trained by \textit{self-supervised learning} on all VG images (without labels), and the relational reasoning component is trained on the IM and SGs (1\% or 10\%) from the parallel set, and fine-tuned from the \textit{ground truth graphs} (99\% or 90\%), instead of the text-to-graph predictions.

\item\textbf{FSPB}: Here, both the backbone and the relational reasoning component are trained by \textit{supervised learning} on $100\%$ of the VG annotated images. Meaning that we have redefined the parallel set to include $100\%$ of the VG training data and we do not need to substitute the images with the text set anymore. The goal of this setting is to compute the maximum accuracy that our model achieves, when we have all the annotated images with ground truth SGs, instead of using their textual scene descriptions. The results of this settings are not included as a separate bar so that the other bars maintain a meaningful scale. Instead, they are written above each table.
\end{itemize}

We use the Recall@K (\textbf{R@K}) for a metric. R@K computes the mean prediction accuracy in each image given the top $K$ predictions. For the complete set of results under constrained and unconstrained setups~\citep{yu2017visual}, and also with the Macro Recall~\citep{9412945} (\textbf{mR@K}), refer to the supplementary materials. 

Figure \ref{fig_golden_obj} presents the results of the ablation study. As shown, fine-tuning with textual scene descriptions improves the classification results under all settings (TXM), substituting a large proportion of the omitted images. Furthermore, the results even outperform FSPB under PredCls (recall that the scene descriptions are sometimes complementary to image annotations and contain additional information). 

\begin{figure*}[t]
	\begin{center}
		\includegraphics[width=0.95\textwidth]{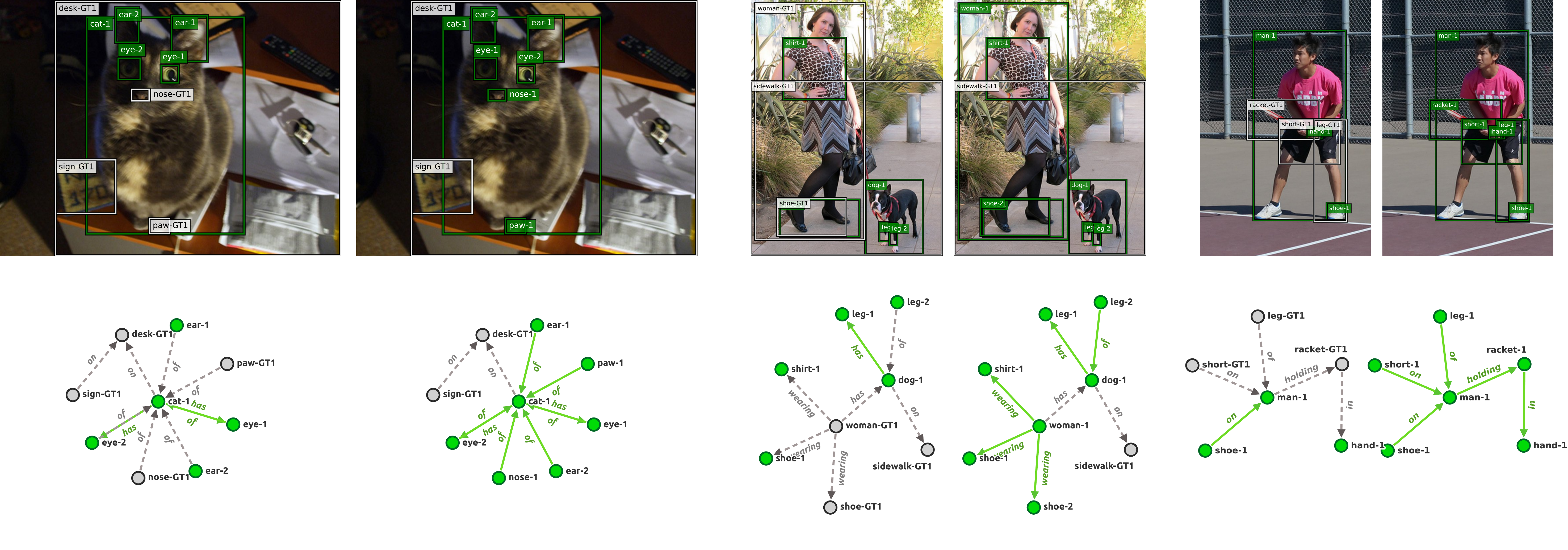}
	\end{center}
	\caption{Qualitative examples of improved scene graph classification results (Recall@100) before and after (from left to right) fine-tuning the model using the knowledge in texts. Green and gray colors indicate true positives and false negatives concluded by the model.
	}	\label{qual}
\end{figure*}

Table \ref{full_results} presents additional results also using different text-to-graph baselines. We can see that fine-tuning with the predicted graphs using T5, is as effective as fine-tuning with the crowd-sourced ground truth graphs (GT), and in some settings even better (object classification with 1\%). Notice that compared to the self-supervised baseline, we gained up to {\raise.12ex\hbox{$\scriptstyle\sim$}}5\% \textit{relative improvement} in object classification, more than {\raise.12ex\hbox{$\scriptstyle\sim$}}26\% in scene graph classification, and {\raise.12ex\hbox{$\scriptstyle\sim$}}31\% in predicate prediction accuracy. As expected, the choice of text-to-graph module has a larger effect on the PredCls compared to the SGCls and ObjCls, due to the fact that SGCls and ObjCls rely heavily on the image-based features, whereas PredCls has a strong dependency to relational knowledge. In supplementary materials we also provide additional results on the improvements per object class after fine-tuning the model with the textual knowledge (From SCH to TXM) and show that most improvements occur in under-represented classes. Figure \ref{qual} provides some qualitative examples of the predicted scene graphs before and after fine-tuning with the texts.

% SOTA TABLE
{\renewcommand{\arraystretch}{1.2}
	\begin{table}
		\centering
		\small
		\setlength{\tabcolsep}{3pt}
        % \resizebox{\columnwidth}{!}
        %\scalebox{1.0\columnwidth}
	    {
			\begin{tabularx}{\columnwidth}{c|cc|cc}
				\hline
				\centering \multirow{2}{*}{Method} & \multicolumn{2}{c|}{SGCls} & \multicolumn{2}{c}{PredCls}  \\
				 & R@50 & R@100 &   R@50 & R@100 \\
				\hline
				\hline
				\centering VRD
				~\citep{lu2016visual}           & $11.8$ & $14.1$ & $27.9$ & $35.0$ \\
				%\centering IMP \citep{xu2017scene}              & $21.7$ & $24.4$ & $44.8$ & $53.0$ \\
				\centering IMP+
				~\citep{xu2017scene}        & $34.6$ & $35.4$ & $59.3$ & $61.3$ \\
				%\centering FREQ \citep{zellers2018neural}       & $32.4$ & $34.0$ & $59.9$ & $64.1$ \\
				\centering SMN
				~\citep{zellers2018neural}     & $35.8$ & $36.5$ & $65.2$ & $67.1$ \\
				\centering KERN
				~\citep{chen2019knowledge}        & $36.7$ & $37.4$ & $65.8$ & $67.6$ \\
				\centering VCTree
				~\citep{tang2019learning}       & $38.1$ & $38.8$ & $66.4$ & $68.1$ \\
				\centering CMAT				~\citep{chen2019counterfactual}
				& $39.0$ & $39.8$ & $66.4$ & $68.1$\\
				\centering SIG
				~\citep{wang2020sketching}  & $36.6$ & $37.3$ & $66.3$ & $68.1$\\
				\centering GB-Net [Zareian et al. 2020]
				%~\citep{zareian2020bridging}
				& $38.0$ & $38.8$ & $66.6$ & $68.2$\\
				%\centering Schemata		%~\citep{sharifzadeh2021classification}
				%& $39.1$ & $39.8$ & $66.9$ & $68.4$ \\
				\hdashline
				\centering \textbf{TXM} & \bm{$39.0$} &\bm{$39.9$} &\bm{$66.7$}&\bm{$68.3$}  
				\\
				\hline
		\end{tabularx}
		}
		\caption{Comparing the general performance of the architecture to some other methods under the VG test set.}
		\label{table_sota}
	\end{table}
}

Note that while a significant stream of research works on the Visual Genome has been focused on utilizing 100\% of the annotated image data and using a pre-trained VGG-16~\citep{simonyan2014very} backbone, in this work, we are focused on the few-shot learning setting and using a ResNet-50~\citep{he2016deep} based self-supervised backbone (BYOL~\citep{grill2020bootstrap}).  Nevertheless, to gain an intuition on our general performance, Table \ref{table_sota} present the results of our architecture using a VGG-16~\citet{simonyan2014very}, outperforming other works.

%Conclusion
\section{Conclusion}\label{sec_conclusion}
In this work, we proposed the first relational image-based classification pipeline that can be fine-tuned directly from the large corpora of unstructured knowledge available in texts. We generated structured graphs from textual input using different rule-based or embedding-based approaches. We then fine-tuned the relational reasoning component of our classification pipeline by employing the canonical representations of each entity in the generated graphs. We showed that we gain a significant improvement in all settings after employing the generated knowledge within the classification pipeline. In most cases, the accuracy was similar to when using the ground truth graphs that are manually annotated by crowd-sourcing.

%References
\bibliography{aaai22}

\cleardoublepage
\appendix

\setcounter{figure}{0}    
\setcounter{table}{0}    
\section{Appendix}

In this section, we present the implementation details and the statistical information of the VG splits that we have used for training and testing. Furthermore, we present more qualitative images and complementary quantitative results (constrained vs unconstrained, mR@K vs R@K). Finally, we present the results \textit{per predicate} and \textit{per object} category.

\begin{figure*}[b] 
	\begin{center}
		\includegraphics[width=0.99\textwidth]{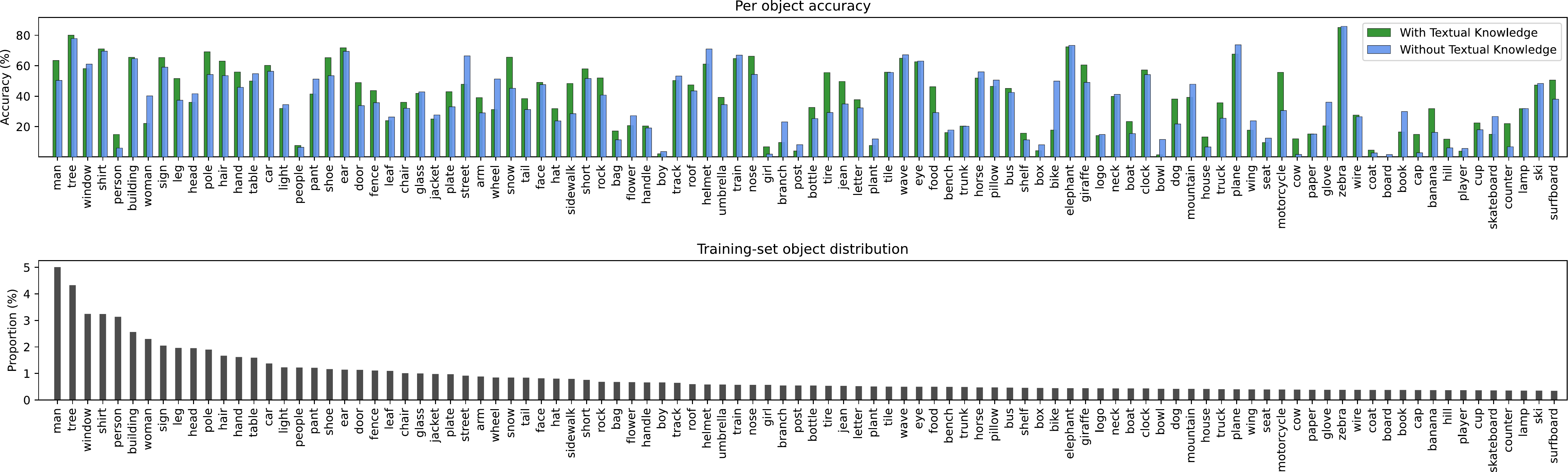}
	\end{center}
	\caption{This plot shows per object classification improvement (Recall@ 100) before and after fine-tuning our model with textual knowledge with 1\% split, the bottom plot shows the sample proportion for the objects in Visual Genome training set.}	\label{fig_obj_acc}
\end{figure*}

\subsection{Implementation Details}
We use ResNet-50~\citep{he2016deep} for the backbone. We train the supervised backbones on the corresponding split of visual genome training set with the Adam optimizer~\citep{kingma2014adam} and a learning rate of $10^{-5}$ for 20 epochs with a batch size of 6. We train the self-supervised backbones with the BYOL~\citep{grill2020bootstrap} approach. We fine-tune the pre-trained \textit{self-supervised} weights over ImageNet~\citep{deng2009imagenet}, on the entire training set of Visual Genome images in a self-supervised manner with no labels, for three epochs with a batch size of 6, SGD optimizer with a learning rate of $6 \times 10^{-5}$, the momentum of 0.9 and weight decay of $4 \times 10^{-4}$. Similar to BYOL, we use an MLP hidden size of 512 and a projection layer size of 128 neurons. Then for each corresponding split, we fine-tune the weights in a supervised manner with the Adam optimizer and a learning rate of $10^{-5}$ for four epochs with a batch size of 6.

After extracting the image-based embeddings from the penultimate fully connected (fc) layer of the backbones, we feed them to an fc-layer with 512 neurons and a Leaky ReLU with a slope of 0.2, together with a dropout rate of 0.8. This gives us initial object node embeddings. We apply an fc-layer with 512 neurons and Leaky ReLU with a slope of 0.2 and dropout rate of 0.1 to the extracted spatial vector $\mathbf{t}$ to initialize predicate embeddings. We take four graph transformer layers of 5 heads for the relational reasoning component, each with 2048 and 512 neurons in each fully connected layer. We initial the layers using Glorot weights~\citep{glorot2010understanding}. We train our supervised models with the Adam optimizer and a learning rate of $10^{-5}$ and a batch size of 22, with five epochs for the $1\%$ and 11 epochs for $10\%$. We train our self-supervised model with the Adam optimizer and a learning rate of $10^{-5}$ and a batch size of 22, with six epochs for the $1\%$ and 11 epochs for $10\%$. Finally, we train our self-supervised model, including the textual knowledge with the Adam optimizer and a learning rate of $10^{-5}$ with a batch size of 16, with six epochs for the $1\%$ and 11 epochs for $10\%$. We use a Geforce RTX 2080 for our experiments.

\subsection{Dataset Details}
We use the sanitized version \cite{xu2017scene} of Visual Genome dataset which contains scene images with their corresponding scene graphs and textual descriptions. We use 57,723 samples as the full (100\%) training set, 5,000 samples as our validation set, and 26,446 samples as our test set. As discussed in the paper, we randomly sample 1\% and 10\% of the entire training data to create the \textit{parallel} set. The text set contains the remaining data (90\% and 99\%). The statistics of the data are shown in Table \ref{table_stats}. Each split (1\% or 10\%) has been sampled randomly four times while keeping the number of images constant. The number of triples and predicates in Table \ref{table_stats} indicate the mean and variance over the four random samples.

{\renewcommand{\arraystretch}{1.35}
	\begin{table}
		\centering
		\small
		\scalebox{0.90}{
			\begin{tabular}{c|c|c|c}
				\hline
                Total number of & 1\% & 10\% & 100\% \\
				\hline
				\hline
				\centering Images       & $577$             & $5772$                & $57723$ \\
				\centering Triples      & $4115.7 \pm 51.0$ & $40382.2 \pm 181.4$   & $405860$ \\
                \centering Objects      & $150$             & $150$                 & $150$ \\
                \centering Predicates   & $45.25 \pm 0.43$  &	$49.50 \pm 0.5$     & $50$ \\ 
                \hline
		\end{tabular}}
		\caption{The statistics of the \textit{parallel} set and the entire training set. The rows for triples and predicates indicate the mean and variance calculated over four different random splits.}
		\label{table_stats}
	\end{table}
}

\subsection{Additional Qualitative Results}
\begin{figure*}[!b]
	\begin{center}
		\includegraphics[width=.9\textwidth]{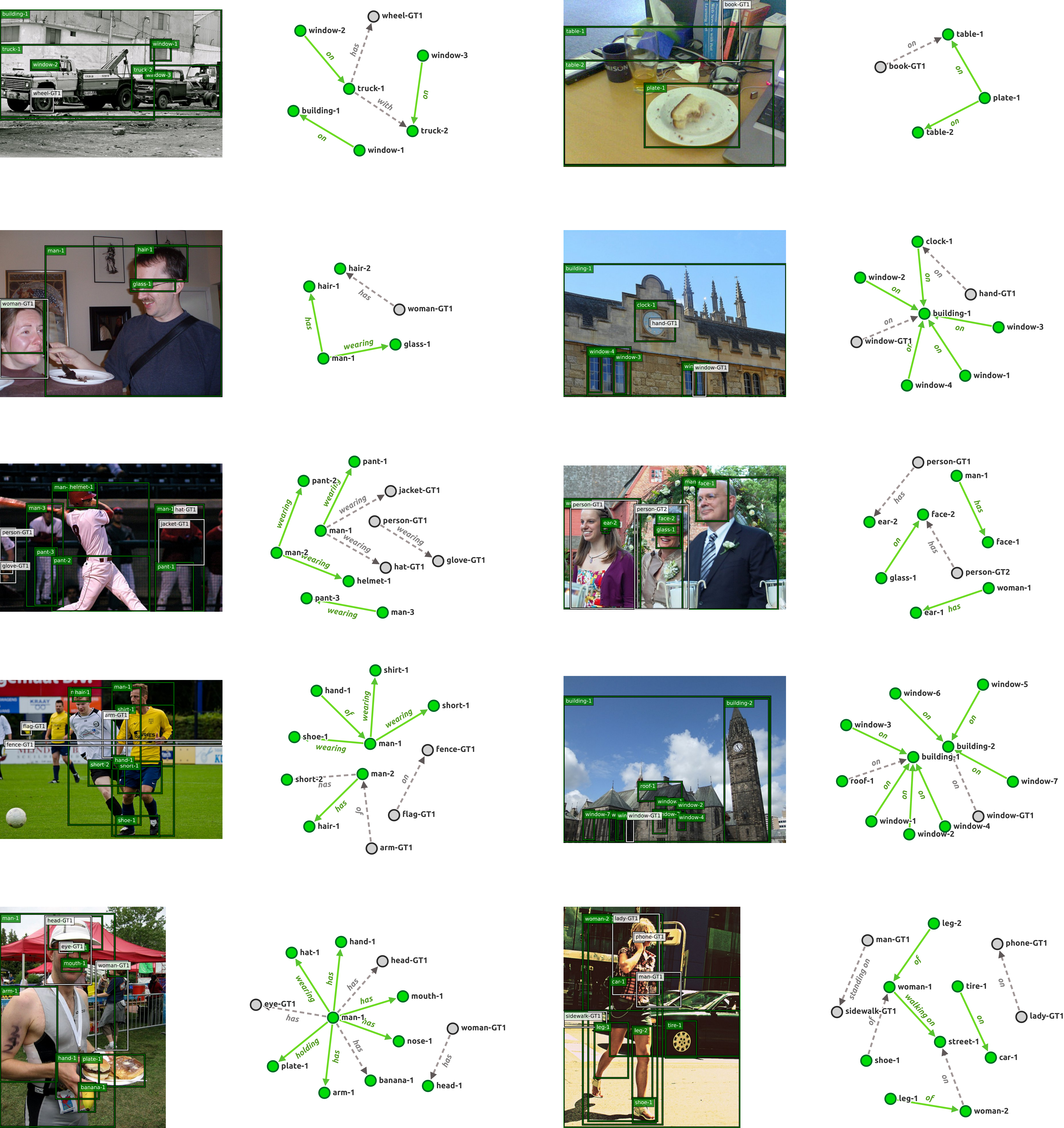}
	\end{center}
	\caption{Qualitative examples of scene graph classification results (Recall@100) using the model trained with the textual knowledge (1\% split). Green and gray colors indicate true positives and false negatives concluded by the model.}	\label{fig_sup_qualitative}
\end{figure*}

Figure \ref{fig_sup_qualitative} present qualitative results of the generated scene graphs from the VG test set and using our model (after fine-tuning it with the textual data).

\subsection{Per Predicate and Per Object Improvements}
\begin{figure*}[b] 
	\begin{center}
		\includegraphics[width=.75\textwidth]{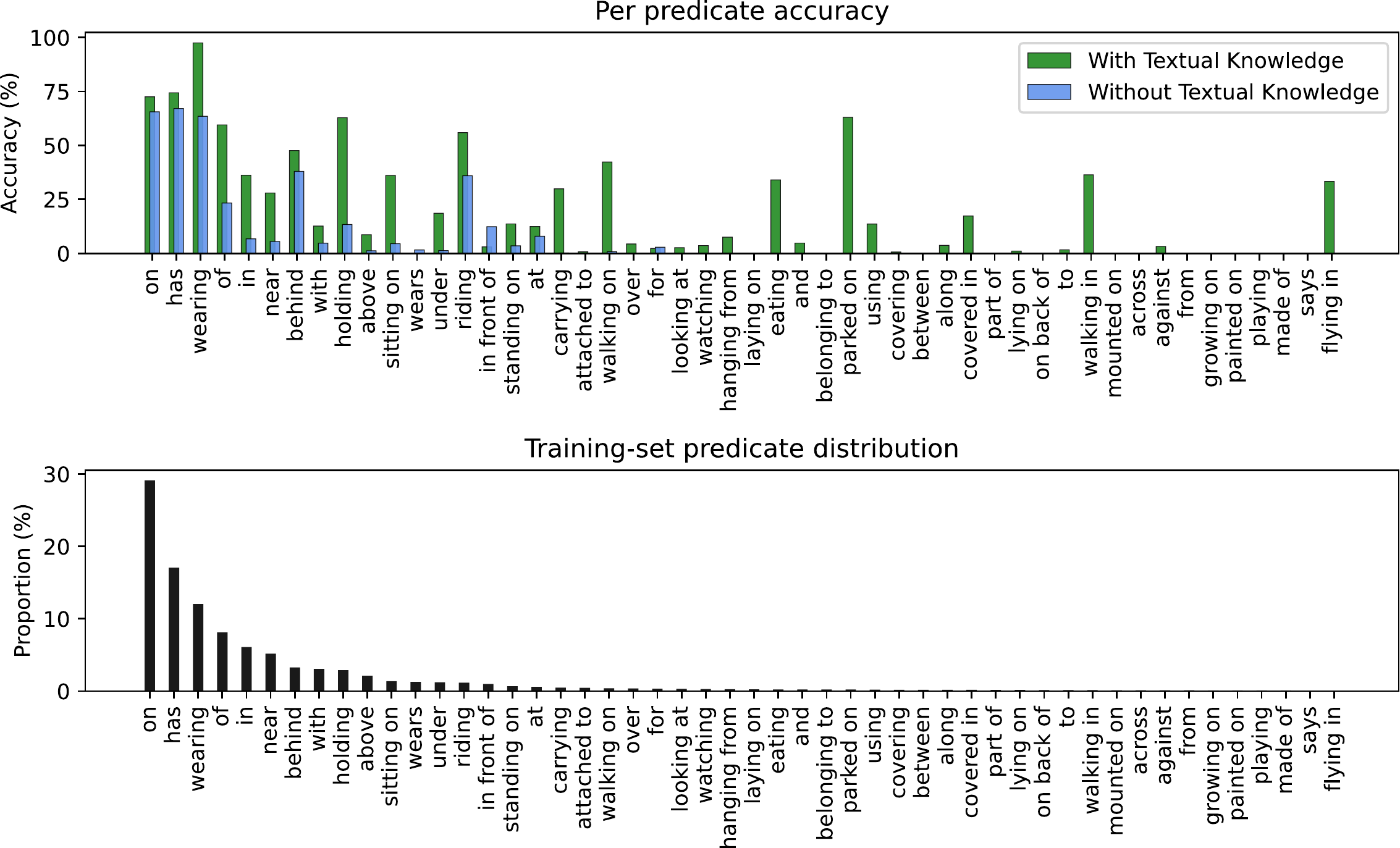}
	\end{center}
	\caption{This plot shows per predicate classification improvement (Recall@ 100) before and after fine-tuning our model with textual knowledge with 1\% split, the bottom plot shows the sample proportion for the predicates in Visual Genome training set.}	\label{fig_sup_per_obj}
\end{figure*}

Figure \ref{fig_obj_acc} and \ref{fig_sup_per_obj} presents bar plots representing per object and per predicate improvement in accuracy, after fine-tuning with the texts. It also presents the frequency of their appearance in the training set. The goal of this table is to understand better where the improvements are happening; the results indicate that most improvements occur in under-represented classes. This means that we have achieved a generalization performance beyond the simple reflection of the dataset's statistical bias. For example, interestingly we have improved the classification of objects such as a \texttt{Motorcycle} and \texttt{Surfboard} even though they only occur a few times in the training set. %and predicates such as 

\subsection{Complementary Quantitative Results}
Table 2-6 present complementary quantitative results including the results under \textit{constrained} and \textit{unconstrained} setups~\citep{yu2017visual}. In the unconstrained setup, we allow for multiple predicate labels, whereas we only take the top-1 predicted predicate label in the constrained setup. Additionally, since the distribution of labeled relations in the Visual Genome is highly imbalanced, we also report the results with the Macro Recall~\citep{9412945,chen2019knowledge} (\textbf{mR@K}) metric; \textbf{mR@K} reflects the improvements in the long tail of the distribution by taking the mean over recall \textit{per predicate}.

{\renewcommand{\arraystretch}{1.2}
	\begin{table*}[!b]
		\centering
		\small
		\scalebox{1.0}{
			\begin{tabular}{c|c|c|cc|cc}
				\hline
				\multirow{2}{*}{} &\centering \multirow{2}{*}{Method} &
				\centering \multirow{2}{*}{Source} &
			     \multicolumn{2}{c|}{R@50} & \multicolumn{2}{c}{R@100}  \\
				& \multirow{2}{*} & & $1{\%}$ & $10{\%}$ & $1{\%}$ & $10{\%}$\\
				\hline
				\hline

				\multirow{4}{*}{\rotatebox[origin=c]{0}{SGCls}} 
                & \centering SPB & \centering Images &  $1.98\pm0.28$ & $16.24\pm1.13$ & $2.40\pm0.30$& $18.15\pm1.23$  \\
				& \centering SCH & \centering Images &  $13.34\pm0.51$ & $30.45\pm0.95$ & $15.75\pm0.64$& $33.94\pm0.94$ \\
				& \centering \textbf{TXM - T5} & \centering \textbf{Images + Text} & $\mathbf{17.14 \pm 0.37}$ & $\mathbf{31.43 \pm 0.46}$ & $\mathbf{19.42 \pm 0.47}$& $\mathbf{35.03 \pm 0.49}$\\
				\cdashline{2-7}
				& \centering {GT} & \centering {Images + GT} & $17.54\pm0.43$ & $31.72\pm0.61$ & $19.59\pm0.49$& $35.33\pm0.65$\\				
				\hline

				\multirow{4}{*}{\rotatebox[origin=c]{0}{PredCls}} 
				& \centering SPB & \centering Images &  $42.46\pm0.99$ & $60.21\pm0.93$ & $55.08\pm1.10$& $71.87\pm0.84$ \\
				& \centering SCH & \centering Images&  $51.77\pm0.59$ & $65.89\pm0.42$ & $63.64\pm0.75$& $76.85\pm0.38$ \\
				& \centering \textbf{TXM - T5} & \centering \textbf{Images + Text} & $\mathbf{68.05\pm0.21}$ & $\mathbf{69.74 \pm 0.20}$ & $\mathbf{79.17 \pm 0.16}$& $\mathbf{80.54 \pm 0.10}$\\
				\cdashline{2-7}
				& \centering {GT} & \centering {Images + GT} &  ${73.89\pm0.19}$ & ${73.25\pm0.23}$ & ${84.30\pm0.16}$& ${83.69\pm0.20}$ \\
				\hline
				
		\end{tabular}}
		\caption{Comparison of R@50 and R@100 with no graph constraints, for SGCls and PredCls tasks.}
		\label{unconst_table_m}
	\end{table*}
}

{\renewcommand{\arraystretch}{1.2}
	\begin{table*}[!t]
		\centering
		\small
		\scalebox{1.0}{
			\begin{tabular}{c|c|c|cc|cc}
				\hline
				\multirow{2}{*}{} &\centering \multirow{2}{*}{Method}&
				\centering \multirow{2}{*}{Source} & \multicolumn{2}{c|}{mR@50} & \multicolumn{2}{c}{mR@100}  \\
				& \multirow{2}{*} & & $1{\%}$ & $10{\%}$ & $1{\%}$ & $10{\%}$\\
				\hline
				\hline

				\multirow{4}{*}{\rotatebox[origin=c]{0}{SGCls}} 
				& \centering SPB &  \centering Images& $0.29\pm0.07$ & $4.65\pm0.62$ & $ 0.43\pm0.10$ & $6.72\pm0.77$   \\				
				& \centering SCH &  \centering Images& $2.54\pm0.42$ & $\mathbf{9.93\pm0.35}$ & $3.71\pm0.49$ & $\mathbf{13.89\pm0.51}$  \\
				& \centering \textbf{TXM - T5} & \centering \textbf{Images + Text} & $\mathbf{4.31 \pm 0.35}$ & ${9.77 \pm 0.35}$ & $\mathbf{6.25 \pm 0.38}$& ${13.75 \pm 0.52}$\\
				\cdashline{2-7}
				& \centering {GT} & 
				\centering {Images + GT}&  ${4.84\pm0.13}$ & ${9.94\pm0.54}$ & ${6.83\pm0.16}$ & ${14.11\pm0.64}$ \\				
				\hline
				
				\multirow{4}{*}{\rotatebox[origin=c]{0}{PredCls}} 
				& \centering SPB & \centering Images&  $5.89\pm0.26$ & $15.02\pm1.37$ & $9.62\pm0.38$ & $23.22\pm1.78$  \\				
				& \centering SCH & \centering Images&  $9.24\pm1.31$ & $20.52\pm0.53$ & $14.28\pm1.63$ & $30.24\pm0.72$  \\				
				& \centering \textbf{TXM - T5} & \centering \textbf{Images + Text} & $\mathbf{23.72 \pm 0.28}$ & $\mathbf{24.37 \pm 0.72}$ & $\mathbf{35.72 \pm 0.37}$& $\mathbf{36.28 \pm 0.80}$\\
				\cdashline{2-7}
				& \centering {GT} & 
				\centering {Images + GT} &  ${27.11\pm0.44}$ & ${25.76\pm0.74}$ & ${40.35\pm0.77}$ & ${38.48\pm0.82}$  \\
				\hline
				
		\end{tabular}}
		\caption{Comparison of mR@50 and mR@100 with no graph constraints, for SGCls and PredCls tasks.}
		\label{unconst_table_mr}
	\end{table*}
}

{\renewcommand{\arraystretch}{1.2}
	\begin{table*}[!t]
		\centering
		\small
		\scalebox{1.0}{
			\begin{tabular}{c|c|c|cc|cc}
				\hline
				\multirow{2}{*}{} &\centering \multirow{2}{*}{Method}&
				\centering \multirow{2}{*}{Source} & \multicolumn{2}{c|}{R@50} & \multicolumn{2}{c}{R@100}  \\
				& \multirow{2}{*} & & $1{\%}$ & $10{\%}$ & $1{\%}$ & $10{\%}$\\
				\hline
				\hline

				\multirow{4}{*}{\rotatebox[origin=c]{0}{SGCls}} 
				& \centering SPB & \centering Images&  $1.65\pm0.26$ & $13.37\pm0.94$ & $1.84\pm0.26$ & $13.90\pm0.97$   \\				
				& \centering SCH & \centering Images& $11.19\pm0.41$ & $25.16\pm0.79$  & $12.12\pm0.47$ & $26.14\pm0.77$ \\
				& \centering \textbf{TXM - T5} & \centering \textbf{Images + Text} & $\mathbf{14.53 \pm 0.34}$ & $\mathbf{26.16 \pm 0.32}$ & $\mathbf{15.28 \pm 0.38}$& $\mathbf{27.22 \pm 0.28}$\\
				\cdashline{2-7}
				& \centering {GT} &
				\centering {Images + GT} &  ${14.72\pm0.38}$ & ${26.33\pm0.45}$ & ${15.36\pm0.38}$ & ${27.37\pm0.47}$ \\
				\hline
				
				\multirow{4}{*}{\rotatebox[origin=c]{0}{PredCls}} 
				& \centering SPB & \centering Images&  $34.92\pm0.81$ & $48.69\pm1.24$ & $40.61\pm0.84$  & $52.51\pm1.19$ \\
				& \centering SCH & \centering Images& $43.13\pm0.59$ & $54.40\pm0.39$  &  $48.10\pm0.54$ & $58.14\pm0.35$  \\        
				& \centering \textbf{TXM - T5} & \centering \textbf{Images + Text} & $\mathbf{58.64 \pm 0.34}$ & $\mathbf{59.31 \pm 0.30}$ & $\mathbf{63.07 \pm 0.37}$& $\mathbf{63.32 \pm 0.24}$\\
				\cdashline{2-7}
				& \centering {GT} &
				\centering {Images + GT}&  ${62.02\pm0.10}$ & ${61.71\pm0.19}$ & ${65.68\pm0.12}$ & ${65.42\pm0.19}$ \\
				\hline
				
		\end{tabular}}
		\caption{Comparison of R@50 and R@100 with graph constraints, for SGCls and PredCls tasks.}
		\label{table_r}
	\end{table*}
}

{\renewcommand{\arraystretch}{1.2}
	\begin{table*}[!t]
		\centering
		\small
		\scalebox{1.0}{
			\begin{tabular}{c|c|c|cc|cc}
				\hline
				\multirow{2}{*}{} &\centering \multirow{2}{*}{Method}&
				\centering \multirow{2}{*}{Source} & \multicolumn{2}{c|}{mR@50} & \multicolumn{2}{c}{mR@100}  \\
				& \multirow{2}{*}	& & $1{\%}$ & $10{\%}$ & $1{\%}$ & $10{\%}$\\
				\hline
				\hline

				\multirow{4}{*}{\rotatebox[origin=c]{0}{SGCls}} 
				& \centering SPB & \centering Images&  $0.19\pm0.04$ & $2.21\pm0.34$ & $0.22\pm0.04$ & $2.43\pm0.36$  \\
				& \centering SCH & \centering Images&  $1.53\pm0.20$ & $\mathbf{5.31\pm0.39}$  &  $1.71\pm0.22$ & $\mathbf{5.80\pm0.40}$ \\
				& \centering \textbf{TXM - T5} & \centering \textbf{Images + Text} & $\mathbf{2.48 \pm 0.26}$ & ${5.04 \pm 0.23}$ & $\mathbf{2.73 \pm 0.29}$& ${5.53 \pm 0.27}$\\
				\cdashline{2-7}
				& \centering {GT} &
				\centering {Images + GT}&  ${2.45\pm0.05}$ & $5.17\pm0.23$    & ${2.68\pm0.06}$ & $5.68\pm0.23$ \\
				\hline
				
				\multirow{4}{*}{\rotatebox[origin=c]{0}{PredCls}} 
				& \centering SPB & \centering Images&  $3.80\pm0.20$ & $8.16\pm0.67$ & $4.56\pm0.22$  & $9.45\pm0.76$ \\
				& \centering SCH & \centering Images& $5.64\pm0.56$ & $11.35\pm0.34$  &  $6.61\pm0.62$ & $12.96\pm0.32$ \\
				& \centering \textbf{TXM - T5} & \centering \textbf{Images + Text} & $\mathbf{15.71 \pm 0.63}$ & $\mathbf{16.20 \pm 0.60}$ & $\mathbf{19.42 \pm 0.84}$& $\mathbf{19.38 \pm 0.48}$\\
				\cdashline{2-7}
				& \centering {GT} &
				\centering {Images + GT}&  ${14.80\pm0.31}$ & ${14.59\pm0.60}$ & ${17.06\pm0.39}$ & ${16.92\pm0.58}$ \\
				\hline
				
		\end{tabular}}
		\caption{Comparison of mR@50 and mR@100 with graph constraints, for SGCls and PredCls tasks.}
		\label{table_m}
	\end{table*}
}

{\renewcommand{\arraystretch}{1.2}
	\begin{table*}[!b]
		\centering
		\small
		\scalebox{1.0}{
			\begin{tabular}{c|c|cc}
				\hline
				\centering \multirow{2}{*}{Method} &
				\centering \multirow{2}{*}{Source}& \multicolumn{2}{c}{Object Classification}  \\
				& \multirow{2}{*} & $1{\%}$ & $10{\%}$ \\
				\hline
				\hline
				\centering SPB &\centering Images&  $14.38\pm0.57$ & $38.45\pm1.21$  \\
				\centering SCH & \centering Images& $40.75\pm0.48$ & $56.97\pm0.76$ \\
				\centering \textbf{TXM - T5} &
				\centering \textbf{Images + Text}&  $\mathbf{42.70 \pm 0.49}$& $\mathbf{58.31 \pm 0.45}$ \\

				\cdashline{1-4}
								
				\centering {GT} &
				\centering {Images + GT}&  ${42.09\pm0.65}$& ${58.60\pm0.56}$ \\

				\hline
		\end{tabular}}
		\caption{Comparison of object classification accuracy (Top-1) on the provided VG splits.}
		\label{table_classification}
	\end{table*}
}

\end{document}